\documentclass[10pt,twocolumn,letterpaper]{article}

\usepackage[pagenumbers]{cvpr} % To force page numbers, e.g. for an arXiv version

\usepackage{graphicx}
\usepackage{amsmath}
\usepackage{amssymb}
\usepackage{booktabs}
\usepackage{multirow}

\usepackage{booktabs, multirow} % for borders and merged ranges
\usepackage{soul}% for underlines
\usepackage[table]{xcolor} % for cell colors
\usepackage{changepage,threeparttable} % for wide tables
\usepackage[pagebackref=true,breaklinks=true,colorlinks,bookmarks=false]{hyperref}

\usepackage{xcolor}
\definecolor{bittersweet}{rgb}{1.0, 0.44, 0.37}
\definecolor{mygreen}{rgb}{0.29, 0.7, 0.48}
\usepackage[capitalize]{cleveref}
\crefname{section}{Sec.}{Secs.}
\Crefname{section}{Section}{Sections}
\Crefname{table}{Table}{Tables}
\crefname{table}{Tab.}{Tabs.}

\newcommand{\myparagraph}[1]{\vspace{0.25em}\noindent\textbf{#1}}
\newcommand{\tablestyle}[2]{\setlength{\tabcolsep}{#1}\renewcommand{\arraystretch}{#2}\centering\footnotesize}
\def\cvprPaperID{5426} % *** Enter the CVPR Paper ID here
\def\confName{CVPR}
\def\confYear{2022}
\def\TaskName{UVLP}
\def\TaskFullName{unsupervised V+L pre-training without parallel data}
\def\ModelName{$\mu$-VLA}
\def\ModelFullName{Unsupervised Vision-and-Language Pre-training via Retrieval-based Multi-Granular Alignment}
\def\uvisualbert{U-VisualBERT}
\newcommand{\head}[1]{\noindent\textbf{#1}}

\begin{document}

%%%%%%%%% TITLE - PLEASE UPDATE
\title{Unsupervised Vision-and-Language Pre-training via Retrieval-based Multi-Granular Alignment} 

\author{
Mingyang Zhou$^1$\thanks{The two authors contribute equally.} \quad Licheng Yu$^3$\footnotemark[1]  \quad  Amanpreet Singh$^3$  \quad  Mengjiao Wang$^3$  \quad  Zhou Yu$^2$  \quad  Ning Zhang$^3$
\\ 
$^1$Uiversity of California, Davis   \quad  $^2$ Columbia University  \\  $^3$Meta AI\\
\tt{\small minzhou@ucdavis.edu}, \tt{\small zy2461@columbia.edu}, \tt{\small\{lichengyu, asg, mengjiaow, ningzhang\}@fb.com} \\
}

\maketitle

%%%%%%%%% ABSTRACT
\begin{abstract}
Vision-and-Language (V+L) pre-training models have achieved tremendous success in recent years on various multi-modal benchmarks.
However, the majority of existing models require pre-training on a large set of parallel image-text data, which is costly to collect, compared to image-only or text-only data.
In this paper, we explore unsupervised Vision-and-Language pre-training (\TaskName) to learn the cross-modal representation from non-parallel image and text datasets. 
We found two key factors that lead to good unsupervised V+L pre-training without parallel data:  $(i)$ \textit{joint image-and-text input} $(ii)$ \textit{overall image-text alignment (even for non-parallel data)}. 
Accordingly, we propose a novel unsupervised V+L pre-training curriculum for non-parallel texts and images. 
We first construct a weakly aligned image-text corpus via a retrieval-based approach, then apply a set of multi-granular alignment pre-training tasks, including region-to-tag, region-to-phrase, and image-to-sentence alignment, to bridge the gap between the two modalities.
A comprehensive ablation study shows each granularity is helpful to learn a stronger pre-trained model.
We adapt our pre-trained model to a set of V+L downstream tasks, including VQA, NLVR2, Visual Entailment, and RefCOCO+. 
Our model achieves the state-of-art performance in all these tasks under the unsupervised setting.
\end{abstract}

%%%%%%%%% BODY TEXT
\begin{figure}[t!]
\centering
\includegraphics[width=0.9\linewidth]{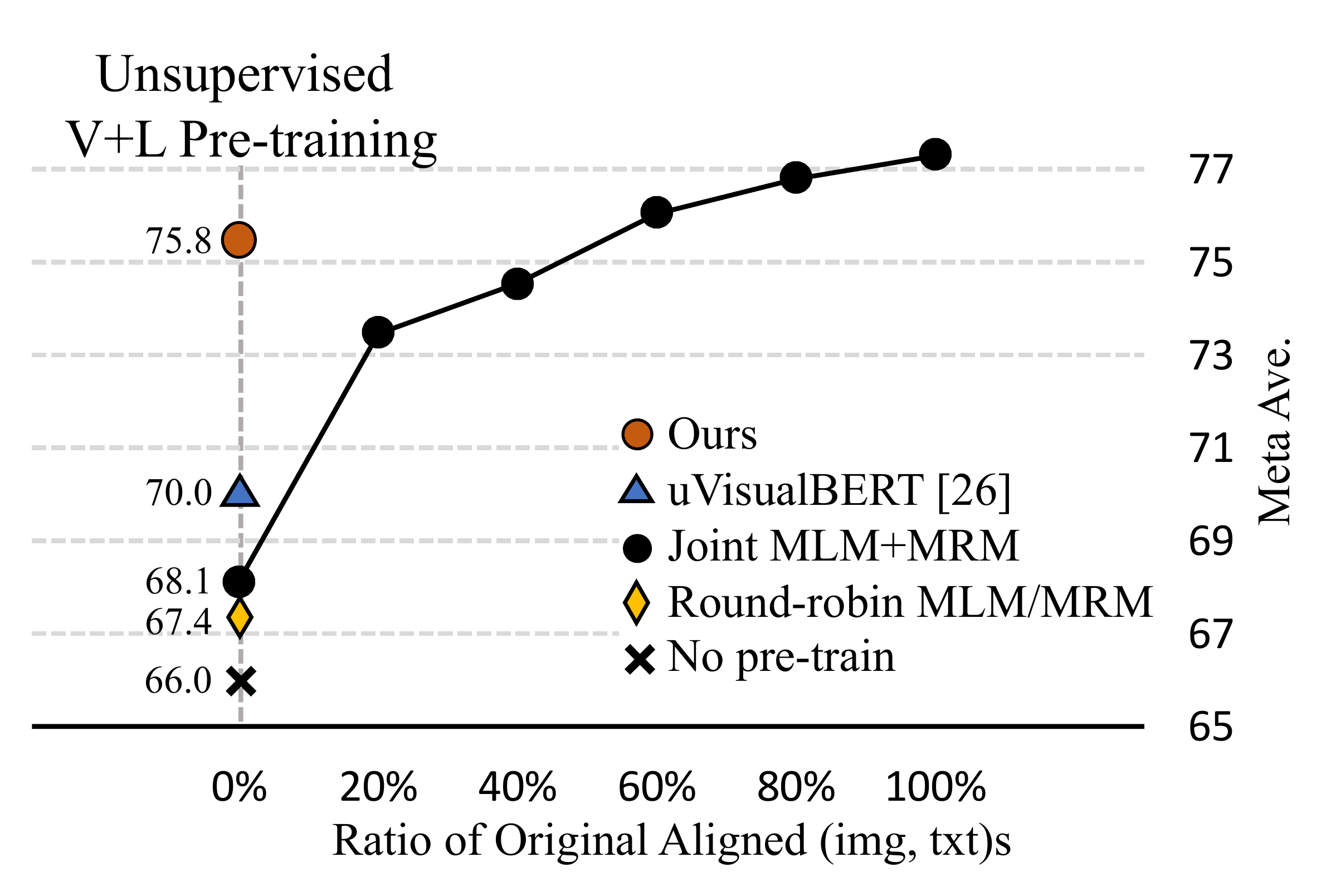}
\caption{Meta average scores of VQA, NLVR2, VE, and RefCOCO+ fine-tuned from different pre-trained models. All pre-training are conducted on Conceptual Captions (CC) with different ratio of parallel data, i.e., a fixed amount of data is originally aligned while the rest is randomly shuffled. 0\% refers to the case of unsupervised V+L pre-training. We also plot the performance of our proposed approach against \uvisualbert \cite{li2020unsupervised}.
Breakdown of the accuracy of each task is listed in the supplementary file.
} 
\label{fig:intro}
\end{figure}

\section{Introduction}\label{section:intro}

% why we need unpaired pre-training
Vision-and-Language pre-trained (VLP) models~\cite{lu202012,chen2020uniter,lu2019vilbert,tan2019lxmert,yu2020ernie,li2020unicoder,su2019vl,kim2021vilt,huang2020pixel,wang2021simvlm,li2020unimo,li2021albef,huang2021soho} that learn the joint cross-modal representation have revolutionized the research on various vision-and-language tasks in recent years.
However, the success of VLP models relies on the availability of a large-scale aligned image-text corpora. 
The widely used crowd-sourced pre-training datasets such as MS COCO \cite{lin2014microsoft, chen2015microsoft} and Visual Genome \cite{krishna2017visualgenome} require expensive human annotations which are hard to scale up. 
Recently, the web crawled image-text datasets like Conceptual Captions 3M~\cite{sharma2018conceptual} and CC12M~\cite{changpinyo2021cc12m}, and SBU Captions~\cite{ordonez2011sbu}, \etc have dramatically reduced the need for massive human annotation but still require heavy post-cleaning procedures to get aligned image-text pairs.
In comparison, the language corpora and image collection are readily available from the web.
The convenience of getting a large-scale single-modality data has benefited the self-supervised learning of vision~\cite{bao2021beit,moco,simclr} and language~\cite{devlin2018bert,roberta} domains respectively.  
This raises a question: Can we take advantage of easily-accessible large-scale single-modality data to perform unsupervised V+L pre-training without parallel text and images (\TaskName)?

We define \TaskName~as follows: 
given the separately crawled image collection $\mathbf I = \{\mathbf i_1, \mathbf i_2, \dots, \mathbf i_{n^I}\}$ and text corpus $\mathbf T = \{\mathbf t_1, \mathbf t_2, \dots, \mathbf t_{n^T}\}$, we aim to pre-train a multi-modal model from such data.
\uvisualbert \cite{li2020unsupervised} is the first \TaskName~work, where the authors have trained their model on un-aligned text and image data in a round-robin fashion and simply use object tags as an anchor point to bridge the gap between the two modalities. 
Their research demonstrates that a shared multi-modal embedding can be learned by just presenting a single modality at a time.
This however introduces an input discrepancy between pre-training and fine-tuning stages as each downstream V+L task requires both modalities (image, text) as the input. 
In this work, we investigate ($i$) whether presenting a joint image-text data from non-parallel data  would improve the learned joint embedding space.
Furthermore, ($ii$) if joint image-text data is fed into the model, how does its latent alignment affect the cross-modal representation learning? 

To explore these two questions, we simply use the images and captions from Conceptual Captions (CC) dataset \cite{sharma2018conceptual} as independently collected uni-modal corpus and perform the following analysis.
First, we compare the pre-trained model's performance between the two data input strategies:
one is presenting one image or text at a time (round-robin) and the other is presenting a concatenation of a pair of randomly sampled image and text (0\% alignment ratio). 
Second, we prepare five sets of image-text pairs from Conceptual Captions with different levels of pairwise alignment by controlling the ratio of original aligned image-text data from 20\% to 100\% (while the remaining is randomly sampled from each modality). 
A single-stream transformer is used for all experiments with the standard pre-training objectives: masked language modeling (MLM) on language input and masked region modeling (MRM) on vision input. After pre-training, we adapt the model to a series of four downstream V+L tasks, including VQA~\cite{antol2015vqa}, NLVR2~\cite{suhr2018corpus}, Visual Entailment (VE)~\cite{xie2019visual}, and RefCOCO+~\cite{yu2016modeling}. 
The performance is measured as the meta average of all tasks after fine-tuning. 
The results are summarized in Fig.~\ref{fig:intro}. 
From Fig.~\ref{fig:intro}, it is clear that joint MLM+MRM learning outperforms
%brings a higher meta average of 68.1 than the 67.4 from 
round-robin MLM/MRM. 
Such gains show that \textbf{joint image-and-text input is necessary for \TaskName}~even when the input is un-aligned. 
We also observe a strong positive correlation between the alignment of image-text pairs and the meta average of the fine-tuned downstream tasks of the resulting model. 
This conveys a seemingly intuitive but quite important message that \textbf{the more aligned the image-text data is the better the pre-trained model performs}. 

Inspired by these analyses, we propose \ModelFullName~(\ModelName), which uses our novel unsupervised V+L pre-training curriculum for non-parallel data.
We first construct a weakly-aligned image-text dataset via retrieval.
% construction and let the model gradually learn a multi-granularity alignment during the pre-training.
Given an image, we take its detected object tags as the reference sentence and retrieve the closest sentences from the text corpus via sentence BERT embedding~\cite{reimers-gurevych-2019-sentence} similarity.
Though the constructed pairs are noisy, the mere weak alignment of concepts is key to learning the latent alignment.
We propose to let the model gradually learn a multi-granular alignment, i.e., region-to-object tag level, region-to-noun phrase level, and image-to-sentence level to more effectively bridge the gap between the two modalities.
We show how each granularity learned from the weakly-aligned pairs contributes to the final pre-trained model's performance.
Experiments show our approach achieves the state-of-art performance
%meta average of 75.5\% 
(in Fig.~\ref{fig:intro}), with a clear gain over~\cite{li2020unsupervised} on the 4 downstream tasks.

Towards practical applications, we also validate the effectiveness of our approach under a more realistic setting, where the images are from CC and the captions are from BookCorpus (BC)~\cite{Zhu_2015_ICCV}.
Similar performance gains are achieved in this harder setting, showing the robustness of our approach.

To summarize, our contributions are three-fold: 
(\textit{i}) We analyze what leads to a good unsupervised V+L pre-training and found two key factors: joint image-and-text input, and overall alignment between image-text pairs.
(\textit{ii}) Accordingly, we propose a novel retrieval-based pre-training curriculum, which applies multi-granular alignment pre-training tasks between weakly aligned image-text pairs to bridge the gap between the two modalities.
(\textit{iii}) We provide comprehensive experiments and analyses showing the robustness of our approach when compared to SOTA supervised and unsupervised V+L pre-training methods.

\section{Related Work}
\begin{figure*}[ht!]
\centering
\includegraphics[width=15.4cm]{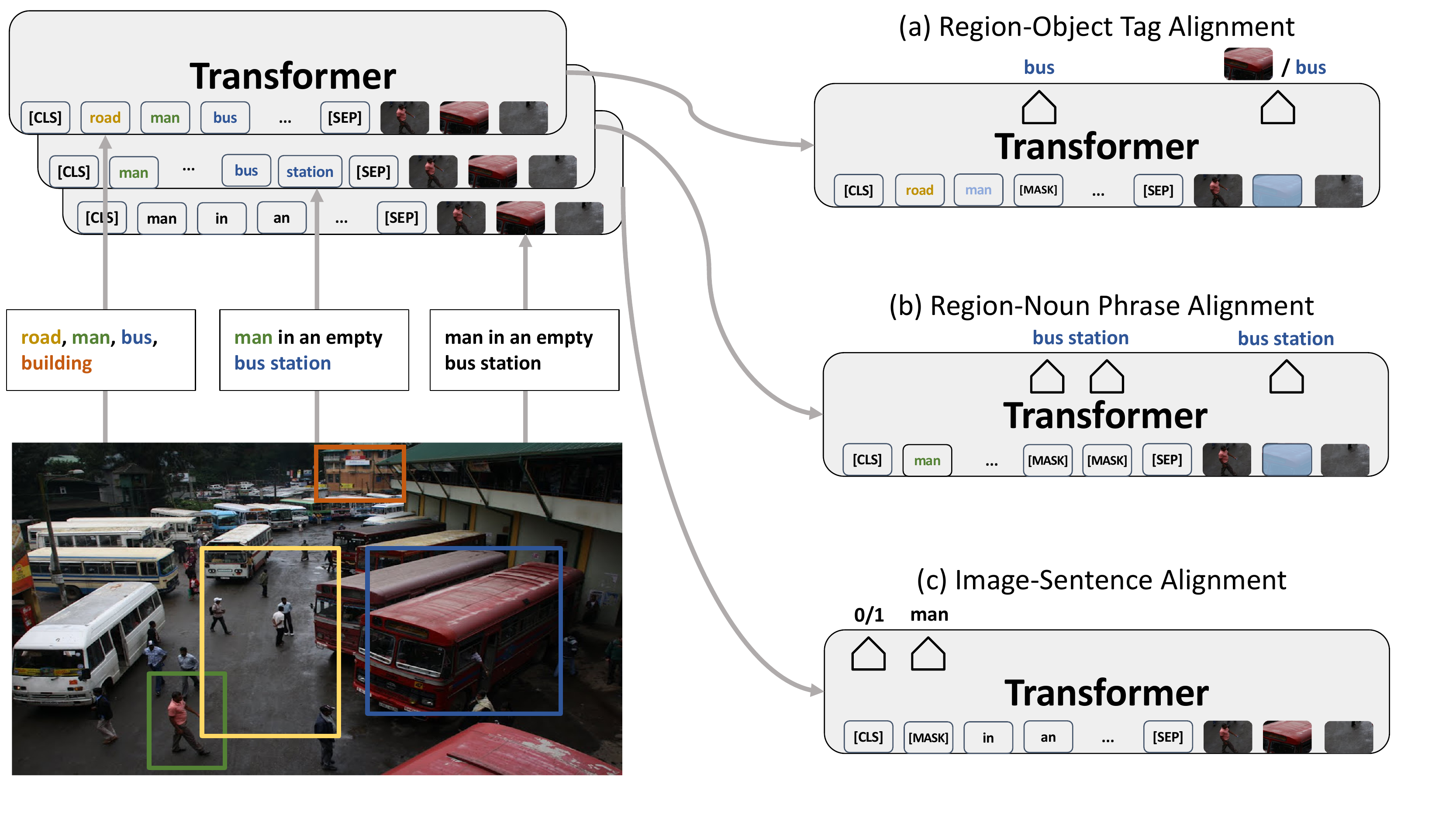}
\caption{Overview of our method. On the left we form three types of image-text pairs as input data to learn cross-modal alignment on three different granularities: region-tag alignment, region-phrase alignment, and image-text alignment. The models is iteratively pre-trained on each granularity and the model parameters are shared. On the right-hand side, we demonstrate the details of the pre-training objectives for each granularity.  
}
\label{fig:model}
\end{figure*}

\paragraph{Vision-and-Language Pre-training}  
Inspired by the success of natural language processing~\cite{devlin2018bert, brown2020language}, there is a recent surge of interest in pre-training for vision and language.
For example, there are different architectures (\eg two-stream models~\cite{lu2019vilbert,tan2019lxmert,lu202012,yu2020ernie,li2021albef,kamath2021mdetr} vs. single-stream models~\cite{li2019visualbert,li2020unicoder,su2019vl,chen2020uniter,li2020unimo}), features (\eg regions~\cite{anderson2018bottom} \vs grids~\cite{huang2020pixel}), backbones (\eg ConvNets~\cite{huang2020pixel} \vs Transformers~\cite{kim2021vilt}) \etc.
All these works aim to exploit the large-scale aligned image-text corpora~\cite{lin2014microsoft,krishna2017visualgenome,sharma2018conceptual,ordonez2011sbu,kamath2021mdetr} to pre-train a powerful multi-modal model, which is then adapted to various downstream V+L tasks, such as VQA~\cite{antol2015vqa}, NLVR2~\cite{suhr2018corpus}, Visual Entailment (VE)~\cite{xie2019visual}, Referring Expression Comprehension~\cite{yu2016modeling}, and Image-Text Retrieval.

Various pre-training tasks have been introduced to achieve this, including the most notable Masked Language Modeling (MLM), Masked Region Modeling (MRM), and Image-Text Matching (ITM).
Several other variants have also been explored, such as predicting the object tags~\cite{li2020oscar, hu2020vivo}, sequence generation~\cite{zhou2020unified, wang2021simvlm}, word-region alignment~\cite{chen2020uniter}.
In this paper, we propose learning a multi-granular alignment between word and region, phrase and region, and image and sentence to better bridge the gap between vision and language.

\paragraph{Unsupervised Vision-and-Language Pre-training without Parallel Data}
Inspired by the works on multi-lingual contextual language modeling~\cite{conneau2019unsupervised,lample2018phrase,lample2017unsupervised, conneau2017word}, \uvisualbert~\cite{li2020unsupervised} first propose the \textit{unsupervised} vision-and-language pre-training without parallel data (\TaskName). 
\uvisualbert~\cite{li2020unsupervised} conducts the masked prediction on the text-only and image-only corpora and introduce the object tags as anchor points to bridge the two modalities.
The authors treat the tags as a sentence when performing MLM, where tags provide alignment with the regions in a picture and implicitly learn a tag-region-level alignment.
However, the anchor tags are still quite different from the text input, missing the sentence completeness and naturalness.
Besides, the latent cross-modal alignment is shown to be important in our analysis (from Fig.~\ref{fig:intro}).
As comparison, our pre-training involves a retrieval-based weakly aligned V+L data construction and learns a more comprehensive multi-granular cross-modal alignment.
With same data as \uvisualbert, our approach achieves a clear and consistent gain across all the downstream tasks in our experiments.

%\section{\ModelFullName}
\section{Method}
In this section, we introduce the two core components of our \ModelName's architecture for~\TaskFullName: 
(1) construct a weakly aligned image-text corpus from independent vision and language data sources; 
(2) our novel pre-training curriculum to enable the model to capture the cross-modal alignment on three granularity including region-to-tag level alignment (RT),  region-to-noun phrase level alignment (RN), and image-to-sentence level alignment (IS). 

\subsection{Model Overview}
We use the well-known single-stream model architecture for our experiments as~\cite{li2019visualbert,li2020unicoder,su2019vl,chen2020uniter,li2020unimo}.
As shown in Fig.~\ref{fig:model}, our main backbone is a single transformer, where we feed the concatenation of visual embeddings of an image and the tokens of a caption as its input.
Given an image $\mathbf i$, we first use an off-the-shelf Faster R-CNN (VinVL~\cite{Zhang_2021_CVPR}) to detect the objects $\mathbf v = \{ v_1, ..., v_{k^v} \}$.
The visual embedding of each region is then encoded as the sum of its regional feature, its location embedding\footnote{The 5-dimensional vector [$\frac{x_1}{W}$,$\frac{y_1}{H}$,$\frac{x_2}{W}$,$\frac{y_2}{H}$,$\frac{(y_2-y_1)(x_2-x_1)}{W.H}$] is projected to the visual embedding space. $(x_1,y_1), (x_2,y_2)$ are the coordinates of the top left and bottom right point of the detected region, and $W,H$ are the image width and height.}, and the modality embedding.
For a given caption $\mathbf t$, we denote its tokenized sequence as $\mathbf t = \{ t_1, ..., t_{k^t} \} $.
After multiple layers of self-attention, the two modalities are fused together and the output hidden vectors can be used for various pre-training tasks.

\subsection{Weakly-aligned Image-Text Corpus}
\label{section:data_aug}
As in the analysis of Sec~\ref{section:intro}, we observe a strong correlation between the degree of image-text alignment in the training data and the performance of the pre-trained model.
% Given the unpaired image collection $I =\{I_1, I_2, \dots, I_{n_I}\}$ and text corpus $T = \{T_1, T_2, \dots, T_{n_T}\}$, 
Inspired by this finding, we believe it important to initialize some weak semantic alignment between the two modalities as the input source.
Specifically, we retrieve $\mathrm{k}$ sentences that are semantically closed to a given $I_i$. 
Previous work~\cite{tan2020vokenization} shows the visually grounded caption covers a good ratio of words that are naturally related to specific visual contents, \eg concrete nouns. 
Thus, we utilize the semantic association between the objects that appear in the image and a candidate sentence as the indicator to measure the alignment degree.

Specifically, we take the object tags $\mathbf o = \{ o_1, ..., o_{k^o} \}$ from the above detected $\mathbf v$ and feed the sequence into an off-the-shelf sentence BERT embedding model~\cite{reimers-gurevych-2019-sentence} to obtain the query embedding $\mathbf e_{\mathbf o}$.
Similarly, we feed each candidate sentence into the same model getting the candidate embedding $\mathbf e_{\mathbf t}$.
We retrieve the top $\mathrm{K}$ candidates with the highest cosine similarity score to form an initial weakly-aligned image-text pairs for a given image $\mathbf i$.
We denote the retrieved captions as $\{ \mathbf t^r (\mathbf i) \}_{r=1}^K$ and the overall weakly aligned corpus as $\mathbf R$.

% Specifically, we filter objects that are detected with confidence score lower than 0.2 and with small bounding box region size that is less than 0.05 of the image size  to focus on critical objects shown in the image. \mingyang{Should appear in the training set up.}
% Specifically, we feed the object-tag sentence and a candidate sentence into the pre-trained off-the-shelf Sentence Bert Embedding model \cite{reimers-gurevych-2019-sentence} to obtain their corresponding embeddings $\mathbf e_{\mathbf{o}}$ and $\mathbf e_{\mathbf{t}}$. 
% We then select the top $\mathrm{k}$ sentence candidates $T_{\text{ret}_i} = {T_{\text{ret}_i}^1, \dots,T_{\text{ret}_i}^k}$ with the highest cosine similarity score with the object list embedding vector $e_{O_i}$
% Eventually, we would have k retrieved sentences $T_{ret} = {T_{ret}^{1}, \dots,T_{ret}^{k}}$ for each image 
% to form the weakly-aligned image-text pairs with the query image $I_i$. 

\subsection{Pre-training Tasks}
% We consider that the success of vision and language pre-training relies on the capability to understand the cross-modal alignment on various granularity including: region-to-object tag level, region-to-noun phrase level, and image-to-sentence level. 
In this subsection, we introduce a set of pre-training objectives that we designed to facilitate the model to capture the different levels of vision and language alignment.
Fig.~\ref{fig:model} shows the overview of our model and its pre-training tasks.

% Unlike \uvisualbert \cite{li2020unsupervised} that focus on just capturing the region to object-tag alignment, we consider the successful unsupervised V+L pre-training should understand the cross-modal alignment on various granularity. Thus, we propose pre-trianing cu

\subsubsection{Region-Tag Alignment Learning}
We first propose to align the object tags onto the image regions.
As shown in Fig.~\ref{fig:model}(a), We concatenate the object tags detected from each image with its source image to form an input pair $[\mathbf o, \mathbf v]$ fed into the model. 
% The detected tags are processed similarly as the normal caption as a sequence of tag tokens $o= [o_{1:l}]$. 
%Following \cite{li2020unsupervised}, the position embedding for each tag token is the spatial box coordinate embedding of its corresponding region. 
% The position embeddings allow the model to distinguish the tags from different regions. 
% Given the pair of object list and image [$o$, $v$] from the training dataset $D$, 
% We randomly mask some tag tokens $o_k$, and some regions $v_j$, and train our model to predict the masked tag tokens and the properties of the masked regions. 
We denote the mask indices as $\mathbf{m}\in \mathbb{N}^M$\footnote{$\mathbb{N}$ is the natural numbers, $M$ is the vocabulary size, and $\mathbf{m}$ is the set of masked indices.}. 
We randomly mask out the object tags and regions, and apply masked language modeling (MLM) and masked region modeling (MRM) for the pre-training.

Specifically, MLM on the object tags is formulated as
\begin{equation*}
    \mathcal{L}_{\text{MLM}}^{\text{R-T}} = - \mathbb{E}_{(\mathbf o, \mathbf v)\sim \mathbf I} \log{P(\mathbf o_{\mathbf m} |\mathbf o_{\backslash \mathbf m}, \mathbf v)},
\end{equation*}
where the goal is to predict the masked object tags based on the observation of their surrounding tags $\mathbf o_{\backslash \mathbf m}$ and image regions $\mathbf v$.
On the vision side, MRM includes both masked region classification loss (MRC) and masked region feature regression loss (MRFR):
\begin{equation*}
    \begin{split}
    \mathcal{L}_{\text{MRM}}^{\text{R-T}} = \mathbb{E}_{(\mathbf o, \mathbf v)\sim \mathbf I}  [f_{\text{MRC}}(\mathbf v_{\mathbf m} | \mathbf v_{\backslash \mathbf m}, \mathbf o) + f_{\text{MRFR}}(\mathbf v_{\mathbf m} | \mathbf v_{\backslash \mathbf m}, \mathbf o) ].
    \end{split}
\end{equation*}
% To calculate the $\mathcal{L}_{\text{MRC}}$ and $\mathcal{L}_{\text{MRFR}}$, we first obtain the transformer output $h_j$ of the masked region $i=v_j$ at the final layer. For $\text{MRC}$, a fully connected (FC) layer $\phi_{\text{MRC}}$ is applied to predict the object category as a normalized distribution over the total number of $K$ classes of the object categories. 
% Thus, $\mathcal{f}_{\text{MRC}}=CE(\phi_{\text{MRC}(h_j)}, c_j)$ is the standard cross-entropy loss. 
% Additionally, for $\text{MRFR}$ we have another FC layer $\phi_{\text{MRFR}}$ to project $h_j$ into the same dimension space of the ROI feature of the masked region $f_j$. Then we apply L2 regression to compute the loss: $f_{\text{MRFR}}=||\phi_{\text{MRFR}(h_j)}- f_j||_2^2$.
Between the two, MRC learns to predict the object semantic class for each masked region  $c(\mathbf v_{\mathbf m})$.
We feed the last hidden output of the masked region $\mathbf v_{\mathbf m}$ into a FC layer and softmax function to predict the classification probabilities $g_{\theta} ( \mathbf v_{\mathbf m} )$.
The objective is to minimize the cross-entropy of
$ f_{\text{MRC}}(\mathbf v_{\mathbf m} | \mathbf v_{\backslash \mathbf m}, \mathbf o) = \mbox{CE}( c(\mathbf v_{\mathbf m}) , g_{\theta} ( \mathbf v_{\mathbf m} ) ) $.
MRFR learns to regress the transformer output of each masked region $\mathbf v_{\mathbf m}$ to its visual features. 
We apply a FC layer to convert its hidden output to a vector $h_\theta (\mathbf v_{\mathbf m})$ of the same dimension as the input regional feature $r(\mathbf v_{\mathbf m})$.
We apply L2 regression:
$f_{\text{MRFR}}(\mathbf v_{\mathbf m} | \mathbf v_{\backslash \mathbf m}, \mathbf o)  = || h_\theta (\mathbf v_{\mathbf m}) - r(\mathbf v_{\mathbf m}) ||^2_2$.

For region-tag alignment learning, we have our pretraining objective function as 
\begin{equation}\nonumber
\mathcal{L}^{\text{R-T}} =  \mathcal{L}_{\text{MLM}}^{\text{R-T}} + \mathcal{L}_{\text{MRM}}^{\text{R-T}} 
\end{equation}

\subsubsection{Region-Noun Phrase Alignment Learning}

Due to the small vocabulary size of object tags, the region-tag alignment learning can only capture a limited amount of localized concepts.
To increase the diversity of concepts, we propose to align the noun phrases from the retrieved sentences to the corresponding regions as well.
As in Fig.~\ref{fig:model}(b), given an image $\mathbf i$ and its retrieved weakly aligned caption $\mathbf t^r (\mathbf i)$, we first detect the noun phrases from the caption using spacy~\cite{spacy2}.
Note the detected noun phrases sometimes contain the attribute words, which further benefits this pre-training task.
We link the noun phrase to its closest visual region by computing the word2vec similarity between the phrase and object tag (associated to each region).
The pre-training still consists of MLM and MRM but are performed with different masking strategy and supervision signal.

Specifically, for both MRM and MLM, we only mask the linked noun phrases from the caption or the linked object regions.
We make the masking probability proportional to the linked similarity score.
Each time we only mask out one modality (phrase or region) to encourage it to be recovered by its linked content.
The region-to-phrase MLM is then formulated as
$\mathcal{L}_{\text{MLM}}^{\text{R-P}} = - \mathbb{E}_{(\mathbf v, \mathbf t^r)\sim \mathbf R} \log{P(\mathbf t^r_{\mathbf m} |\mathbf t^r_{\backslash \mathbf m}, \mathbf v)}$.

On the vision side, we propose using the phrase-guided masked region-to-token classification (p-MRTC) on the masked regions:
\begin{equation*}
    \begin{split}
    \mathcal{L}_{\text{MRM}}^{\text{R-P}} = \mathbb{E}_{(\mathbf v, \mathbf t^r)\sim \mathbf R}  f_{\text{p-MRTC}}(\mathbf v_{\mathbf m} | \mathbf v_{\backslash \mathbf m}, \mathbf t^r),
    \end{split}
\end{equation*}
where we directly classify the masked region to its linked noun phrase (sub-word tokens) in BERT vocabulary.
Enlarging the vocabulary has shown to be beneficial to MRM~\cite{Zhou_2021_CVPR}.
Our proposed p-MRTC leverages the additional noun-phrase to encourage more diverse local region to language alignment.

For region-noun phrase alignment learning, we have our pretraining objective function as 
\begin{equation}\nonumber
\mathcal{L}^{\text{R-P}} =  \mathcal{L}_{\text{MLM}}^{\text{R-P}} + \mathcal{L}_{\text{MRM}}^{\text{R-P}} 
\end{equation}

\subsubsection{Image-Sentence Alignment Learning}
\label{section:itm}
We apply image-text matching (ITM) objective as the previous supervised V+L pre-training research \cite{chen2020uniter,li2020unicoder} to learn the cross-modal sentence-level alignment. 
As in Fig.~\ref{fig:model}(c), given an input pair [$\mathbf v$, $\mathbf t^r$], the final hidden vector of the special token $\text{[CLS]}$ is fed through a FC layer to output a single score $\mathbf s_{\theta}(\mathbf v, \mathbf t^r)$, which predicts if the given image-text input is a semantically matched pair or not. 
We use the label $y\in \{0,1\}$ to indicate if a retrieved pair is a match.
The training objective for the ITM task is to minimize the binary cross-entropy loss:
$
\mathcal{L}_{\text{ITM}} =\mbox{CE}( y ,s_{\theta}(\mathbf{v}, \mathbf{t}^r) ) 
$.
% \begin{equation*}
%     \mathcal{L}_{\text{ITM}} = -\mathbb{E}_{(\mathbf{v},\mathbf{t^r})\sim \mathbf{R}} [y \log s_{\theta}(\mathbf{v}, \mathbf{t}^r) + (1-y) \log (1-s_{\theta}(\mathbf{v}, \mathbf{t}^r))]
% \end{equation*}
On the language side, we also apply standard MLM to help the model learn to align other language tokens besides noun phrases and object tags to the visual context.
The objective function is then formulated as $\mathcal{L}_{\text{MLM}}^{\text{I-S}} = - \mathbb{E}_{(\mathbf v, \mathbf t^r)\sim \mathbf R} \log{P(\mathbf t^r_{\mathbf m} |\mathbf t^r_{\backslash \mathbf m}, \mathbf v)}$.
The image-sentence level alignment pretraining objective function is
\begin{equation}\nonumber
\mathcal{L}^{\text{I-S}} =  \mathcal{L}_{\text{MLM}}^{\text{I-S}} + \mathcal{L}_{\text{ITM}}
\end{equation}

\subsection{Multi-Granular Pre-training Curriculum}
We propose a multi-granular curriculum to iteratively pre-train the model on the region-to-tag, region-to-noun phrase, and image-to-sentence level. 
According to our findings in Sec.~\ref{section:intro}, learning from image-text pairs with higher degree of cross-modal alignment is beneficial to the performance of unsupervised V+L pre-trained model. 
Therefore, we propose using an estimated image-text alignment score to guide our multi-granular pre-training. 
Specifically, we have an ITM header defined in Sec.~\ref{section:itm} to learn the image-text alignment. 
We also use it to predict matching score as a weight to modulate the input data for each of our retrieval-based pre-training tasks. 
This allows us to provide more importance to relatively more aligned image-text pairs over time to help our model to learn better cross-modal alignment on multiple granularities. 

To train the alignment model's ITM classifier, we use our retrieved corpus $\mathbf{R}$ as positive samples and randomly shuffled pairs as negative samples in the first $m$ epochs. This warms up the models to make reasonable estimations on the alignment of image-text input pairs.
After $m$ epochs, we start to incorporate the alignment prediction score $w_{\text{ITM}}$ in our training objective. 
To summarize, our multi-granular pre-training loss is 
\begin{equation}\nonumber
\mathcal{L}=  
\begin{cases}
      \mathcal{L}^{\text{R-T}} + \mathcal{L}^{\text{R-P}} +\mathcal{L}^{\text{I-S}}   &\text{if epoch} < m\\ 
      \mathcal{L}^{\text{R-T}} + w_{\text{ITM}} (\mathcal{L}^{\text{R-P}}  +  \mathcal{L}^{\text{I-S}}) &\text{if epoch} \geq m, \\
    \end{cases}
\end{equation}
where $\mathcal{L}^{\text{R-T}}$, $\mathcal{L}^{\text{R-P}}$, and $\mathcal{L}^{\text{I-S}}$ are the loss functions for region-tag alignment pre-training, region-noun phrase alignment pre-training, and image-sentense alignment pre-training. We set m as 1 in our final implementation.

\begin{table*}[!ht]\centering
\small
\begin{tabular}{l|c|c|c|ccc|c}\toprule
\multirow{2}{*}{ Model } &VQA2 &NLVR2 &VE & \multicolumn{3}{c|}{RefCOCO+} & \multirow{2}{*}{ Meta-Ave } \\
&Test-Dev &Test-P &Test &Dev &TestA &TestB & \\\cmidrule{1-8}
ViLBERT\cite{lu2019vilbert} &70.6 &- &- &72.3 &78.5 &62.6 & - \\
VL-BERT\cite{su2019vl} &71.2 &- &- &71.6 &77.7 &61.0 & - \\
$\text{UNITER}_{\text{CC}}$\cite{chen2020uniter} &71.2 &- &- &72.5 &79.4 &63.7 & - \\
VisualBERT \cite{li2019visualbert,li2020unsupervised} &70.9 &73.9 &- &73.7 &79.5 &64.5 & - \\
Aligned VLP &\textbf{72.5} &\textbf{75.9} &\textbf{78.7} &\textbf{82.1} &\textbf{86.6} &\textbf{75.0} & \textbf{77.3} \\
\midrule
Base &70.1 &51.2 &73.2 &69.4 &74.8 &60.3 & 65.9 \\
\uvisualbert \cite{li2020unsupervised} &71.8 &53.2 &76.8 &78.2 &83.6 &69.9 & 70.0\\
$\text{\ModelName}_{\text{CC}}$ &\textbf{72.1} &\textbf{73.4} &\textbf{77.3} &\textbf{80.3} &\textbf{85.5} & \textbf{73.7} & \textbf{75.8} \\
$\text{\ModelName}_{\text{BC}}$ &71.2 &67.1 &77.1 &79.7 &85.0 &72.7 & 73.8 \\
\bottomrule
\end{tabular}
\caption{Evaluation results on four V+L downstream tasks. Our model trained with un-aligned data ($\text{\ModelName}_{\text{CC}}$, $\text{\ModelName}_{\text{BC}}$) achieves comparable performance with the supervised model trained with aligned data (Aligned VLP). $\text{\ModelName}_{\text{CC}}$ and $\text{\ModelName}_{\text{BC}}$ also outperform {\uvisualbert } on nearly all tasks.}
\label{tab:main}
\end{table*}

\begin{table*}[!ht]\centering
\small
\begin{tabular}{l|c|c|c|ccc|c}\toprule
\multirow{2}{*}{V+L Alignment} &VQA &NLVR2 &VE &\multicolumn{3}{c|}{RefCOCO+} & \multirow{2}{*}{Meta-Ave}\\
&Test-Dev &Test-P &Test &Dev &TestA &TestB & \\\cmidrule{1-8}
$\text{\ModelName}_{\text{CC}}$ (R-T)  &71.7 &52.0 &75.6 &78.7 &83.3 &70.0 & 69.5\\
$\text{\ModelName}_{\text{CC}}$ (R-N)  &71.4 &69.4 &76.5 &77.4 &81.5 & 68.7 & 73.7 \\
$\text{\ModelName}_{\text{CC}}$ (I-S)  &71.6 &71.5 &76.8 &75.7 &80.3 &67.9 & 73.9\\
$\text{\ModelName}_{\text{CC}}$ (R-T + R-N) &71.9 &72.4 &76.4 &79.3 & 84.5 & 71.7 & 75.0 \\
$\text{\ModelName}_{\text{CC}}$ (R-T + R-N + I-S) &\textbf{72.1} & \textbf{73.4} &\textbf{77.3} &\textbf{80.3} & \textbf{85.0} &\textbf{73.7} & \textbf{75.8}\\
\bottomrule
\end{tabular}
\caption{Effect of cross-modal alignment on the three types of granularities: region-tag alignment(R-T), region-noun phrase alignment(R-N), and image-sentence alignment(I-S)}\label{tab:ablation_align}
\end{table*}

\section{Experiments}
In this section, we provide the detailed experimental set up to evaluate our proposed {\ModelName } against previous supervised and unsupervised VLP models. More specifically, we introduce our pre-training dataset, baselines, and our pre-training setting.

\subsection{Pre-training Datasets}
We prepare the un-aligned data under two different settings: (1) We use images and text separately from Conceptual Captions (CC) \cite{sharma2018conceptual} ignoring the alignment information; (2) We use images from Conceptual Captions (CC) \cite{sharma2018conceptual} and text from BookCorpus (BC) \cite{Zhu_2015_ICCV}. 
Setting (1) sets up a fair comparison with previous supervised methods by keeping the domain and the quality of training data consistent. Our proposed model trained in this setting is called \ModelName$_{CC}$. 
Setting (2) mimics a more realistic challenge where we have large-scale images and text data from different domains, in particular the text sources are not similar to captions of the images. \ModelName$_{BC}$ has been trained in this setting.

As introduced in section \ref{section:data_aug}, for each image we retrieve 5 text data points (captions from CC or sentences from BC) from the text corpus that are semantically similar to the detected objects in the image. 
This creates weakly-aligned image-text pairs for our pre-training models. 

\subsection{Baselines}
We compare the performance of our proposed {\ModelName } to the following baselines: 

\head{Base Model} VisualBERT that is initialized from BERT. It does not undergo any pre-training but is directly fine-tuned on the downstream tasks. 

\head{Supervised Pre-trained Models} Supervised pre-trained VLP models that are trained only on CC, including VILBERT\cite{lu2019vilbert}, VL-BERT\cite{su2019vl}, and UNITER\cite{chen2020uniter}. We also report the numbers on the Supervised VisualBERT implemented in \uvisualbert\cite{li2020unsupervised} that is trained on CC and an additional 2.5 Million text segments from BC. 
For fair comparison with our proposed method, we also introduce the aligned vision-language pre-training model (Aligned VLP) that is pre-trained on the 3M (image, caption) pairs from CC and 3M (image, object tag) pairs. 

\head{Unsupervised Pre-trained Models} 
{\uvisualbert } is pre-trained on individual image or text corpus in a round-robin fashion and captures the cross-modal alignment by using detected object tags as the anchor point. 
For fair comparison, we re-implemented this method to pre-train with the VinVL object features\cite{zhang2021vinvl} and BC. 
\subsection {Training Setup}\label{sec:training_setup}
Our transformer architecture consists of 12 layers of transformer blocks, where each block has 768 hidden units and 12 self-attention heads. 
We initialize the model from $\text{BERT}_{base}$ and pre-train for 20 epochs on their respective pre-training datasets with a batch size of 480. The region features for images are obtained from the pre-trained VinVL object detectors \cite{zhang2021vinvl}. We use Adam optimizer \cite{ADAM} with a linear warm-up for the first 10\% of training steps, and set the peak learning rate as 6e-5. After warm up, a linear-decayed learning-rate scheduler gradually drops the learning rate for the rest of training steps. 
All models were trained on 4 NVIDIA A100 GPUs, with 40GB of memory per GPU using MMF\cite{singh2020mmf}.
The pre-training takes 3 days.
We evaluate our pre-trained models on four downstream tasks: Visual Question Answering (VQA 2.0)\cite{anderson2018bottom}, Natural Language for Visual reasoning\cite{suhr2018corpus} ($\text{NLVR}^2$), Visual Entailment\cite{xie2019visual} (VE), and Referring Expression\cite{yu2016modeling} (RefCOCO+). 
% To validate our proposed sentence-image alignment pre-training, we also conduct a zero-shot evaluation with image-text retrieval task on Flickr30K\cite{young-etal-2014-image}. 
Detailed training settings for each task can be found in our supplementary material. 
\subsection{Experimental Results}
We first compare {\ModelName } to various supervised models that are pre-trained on CC and to the state-of-the-art unsupervised V+L pre-training method, {\uvisualbert } on the four downstream tasks. Besides reporting scores for each individual task, we also compute the meta-average score to reflect the overall performance across all tasks. 
The results are summarized in Table \ref{tab:main}.

\myparagraph{Compared to Base.} It is clear from Table~\ref{tab:main} that both \ModelName$_{CC}$ and \ModelName$_{BC}$ outperform the Base model by a large margin on all benchmarks.

\myparagraph{Compared to Aligned VLP.} It also achieves better performance than existing supervised models like VilBERT\cite{lu2019vilbert}, which is potentially due to the usage of better visual regional features of VinVL~\cite{Zhang_2021_CVPR}. When compared to Aligned VLP, which is trained with the same architecture and visual features, our model is only slightly worse. 
This shows the effectiveness of our proposed pre-training curriculum which can learn comparable universal representation across vision and language as the supervised models without any parallel image-text corpus. 

\myparagraph{Compared to UVLP. }Our {\ModelName } also achieves consistently better performance than the previous UVLP method: \uvisualbert. 
This improvement shows how our proposed cross-modal alignment pre-training curriculum effectively bridges the gap across the two modalities.
In particular, our model outperforms {\uvisualbert } in the task of NLVR2 by more than 20\%. 
As NLVR2 is known to benefit more from image-sentence cross-modal alignment from previous supervised V+L pre-training research \cite{chen2020uniter}, this observation indicates that our model is able to capture the instance-level cross-modal alignment without parallel data. 
When {\ModelName } is trained on BC text and CC images \ie \ModelName$_{BC}$, it still achieves comparable or better performance than {\uvisualbert } except for VQA.  
The slight advantage  {\uvisualbert } has over \ModelName$_{BC}$ in VQA is potentially due the similar style between the VQA text and the pre-trained CC captions. 
However, this does not overshadow the overall better performance of {\ModelName}. 
It shows that our proposed method is more robust than {\uvisualbert } training on the uni-modal datasets collected from separate domains, which makes it more useful in practical settings.    
\begin{figure}[h!]
\centering
\includegraphics[width=0.8\linewidth]{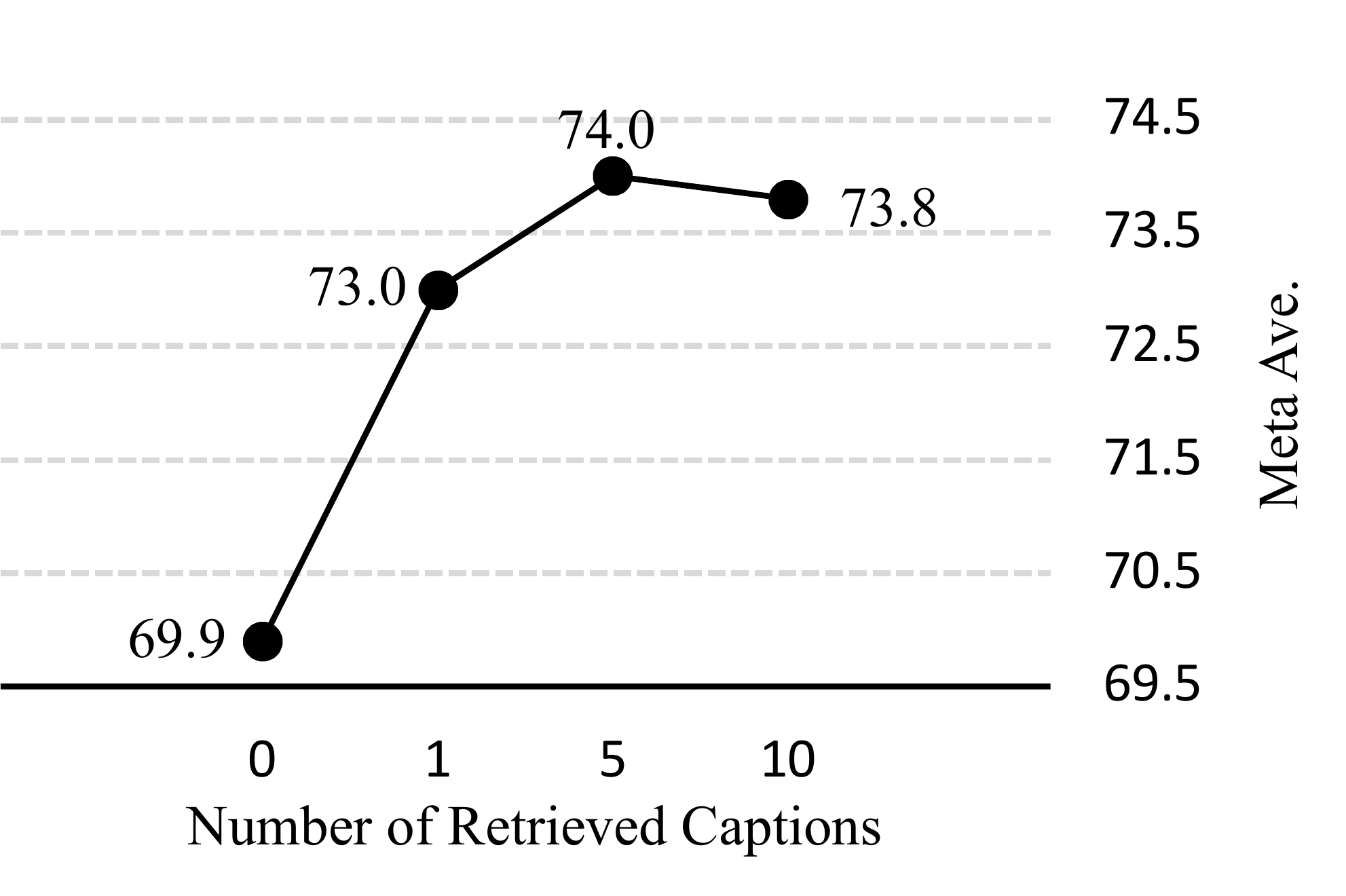}
\caption{Meta average scores of non-parallel V+L pre-training with different number of retrieved candidate sentences.}
\label{fig:ablation_ncandidate}
\end{figure}

\begin{figure*}[h!]
\centering
\includegraphics[width=16cm]{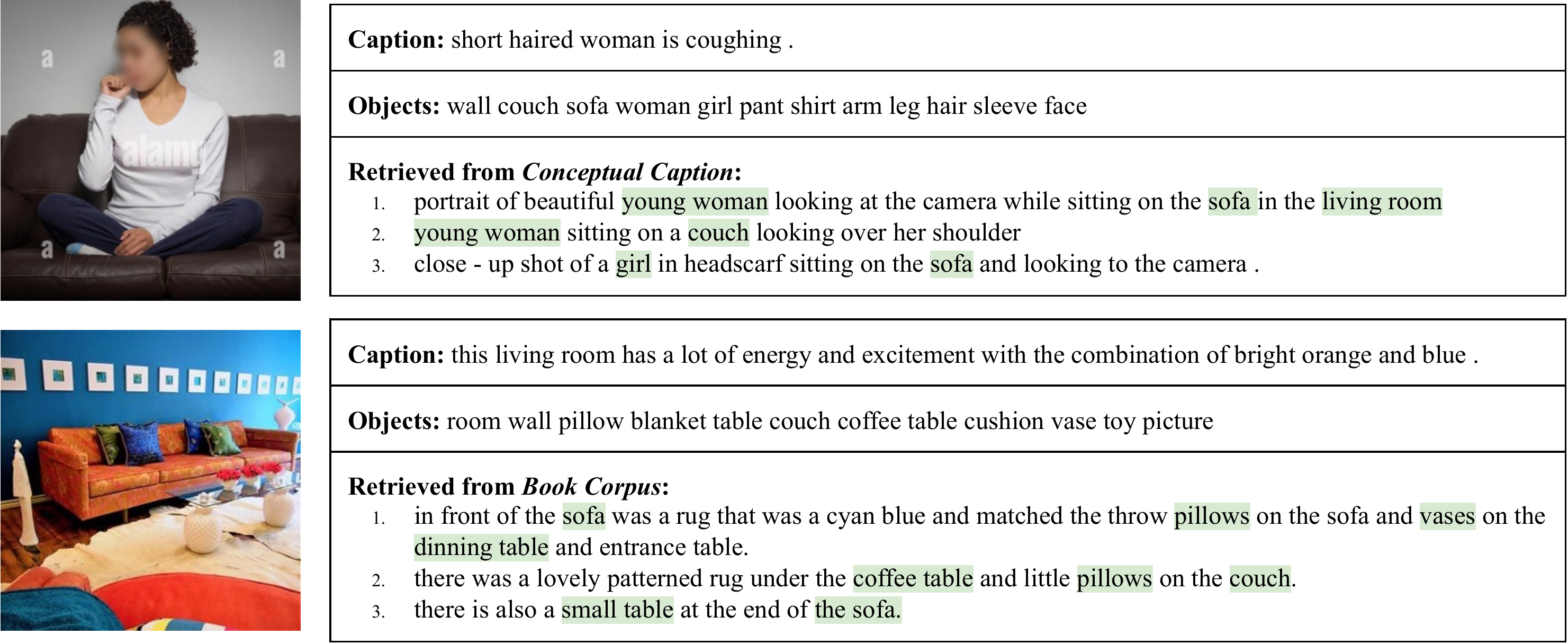}
\caption{Examples of retrieved text from both CC and BC. The covered grounded noun phrases in retrieved sentences are highlighted in green bar for positive examples.}
\label{fig:visualization}
\end{figure*}

\subsubsection{Ablation Study on Multi-Granular Alignment}
We conduct ablation study to verify the effectiveness of the three types of visual-language alignment for unsupervised V+L pre-training, namely region-tag alignment (R-T), region-noun Phrase alignment (R-N), and image-sentence alignment (I-S). We first evaluate each individual type of alignment to measure its usefulness for different downstream tasks. Then, we gradually add each type of alignment into the UVLP.  For this ablation study we pre-train {\ModelName } on CC images and text, and the results are summarized in Table \ref{tab:ablation_align}. 

From Table~\ref{tab:ablation_align}, we can see that aligning local regions to object tags (R-T) and noun phrases (R-N) are especially helpful for the task of RefCOCO+, which requires the model to understand specific objects that natural expressions describe. Meanwhile, aligning the image and sentence at instance-level (I-S) benefits NLVR2 and VE. Especially on NLVR2, the model that captures the global vision and language alignment \ModelName$_{CC}$ (I-S) obtains 19.5\% gain over the model that only learns the local alignments between regions and object tags \ModelName$_{CC}$ (R-T). This observation is consistent with previous research \cite{chen2020uniter}, where the performance of model on NLVR2 is boosted after introducing pre-training objectives that capture the cross-modal alignment in the image-text pairs. Our results demonstrate that even with just weakly-aligned sentences, we can still effectively learn the instance-level cross-modal alignment. 
% \mingyang{Maybe also show zero-shot Image-Sentence Alignment performance to further proves the learned cross-modal alignment on vision and language;} 
Combining the region-tag and region-noun phrase alignment (R-T+R-N) for UVLP, we observe that these two types of grounding and matching compensate each other. \ModelName$_{CC}$ (R-T+R-N) shows a marginal but consistent improvement over models that only learn a single type of local region-to-language alignment (R-T, R-N). After adding object-phrase level alignment we can further improve the performance on NLVR2 and VE, which gives us our best performing model \ModelName$_{CC}$ (R-T + R-N + I-S). 

\subsubsection{Ablation Study on Number of Retrieved Candidates}
We conduct experiments to verify the impact the number of retrieved candidate text for each image has on the performance. We create three variants of pre-training corpus, where the number of retrieved candidate are 1, 5, and 10 based on the rank of the similarity of each candidate text to the query image's detected object tags. The candidate text is sampled from CC. We pre-train our {\ModelName } model with only the pre-training objectives to capture the sentence-image alignment (I-S). For each variant of pre-training corpus, we train the model for the same number of steps. We compute the meta average score for the three resulting pre-trained models and visualize them in Fig.~\ref{fig:ablation_ncandidate}. 

Fig.~\ref{fig:ablation_ncandidate} shows that retrieving more than one candidate text for an image greatly benefits the pre-trained model to learn a better joint representation between vision and language, demonstrated by stronger performance in the downstream tasks. 
We suspect this is because the closeness between the candidate caption and the detected object tags in language embedding space does not always mean a better alignment between the candidate caption and the image. A better and more semantically similar caption candidate for the image could be found in the other caption candidates. However, when we increase the number of candidate captions to 10, we observe a slight drop on the overall performance compared to the model that is pre-trained on corpus with 5 candidate captions. This indicates that having too many candidate captions to form the weakly-aligned pairs with the query image for V+L pre-training may also introduce unnecessary noise. Hence, we set the number of retrieved captions in our experiments to 5.

\subsubsection{Visualization}
To get a sense of the quality of the retrieved sentences, we show some examples of retrieved text from both CC and BC in Fig.~\ref{fig:visualization}. The first row demonstrates a positive case of retrieved captions from CC, where we observe a good coverage of the objects in the image such as ``young woman", ``sofa", and ``couch" in the top retrieved sentences. Similarly, our retrieval method can also retrieve good candidates from BC that describe many visual objects from the image as depicted in row 2. This observation demonstrate the effectiveness of picking candidates based on their closeness to the object list in the language embedding space. 

\begin{figure}[h!]
\centering
\includegraphics[width=0.8\linewidth]{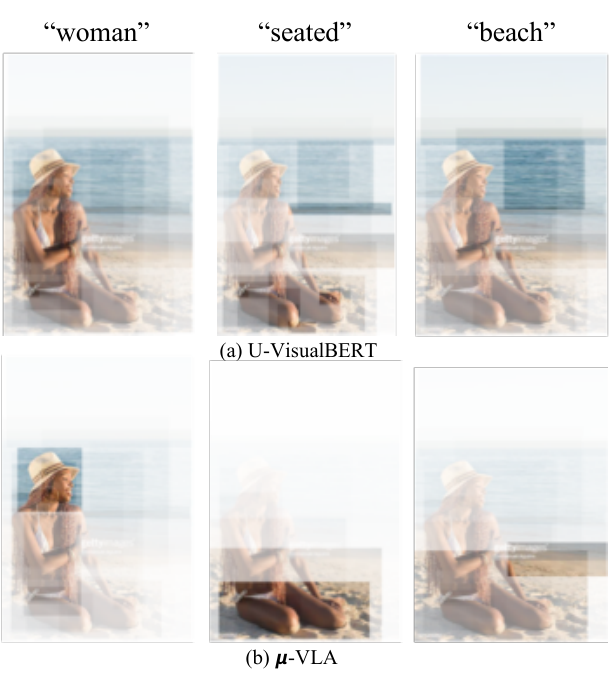}
\vspace{-0.5cm}
\caption{Text-to-image attention given the aligned pair whose caption is ``young woman seated on the beach".}
\label{fig:attention}
\end{figure}

We also compare the text-to-image attention between the pre-trained \uvisualbert~and \ModelName~without task-specific fine-tuning as~\cite{chen2020uniter,Zhou_2021_CVPR}.
As shown in Fig.~\ref{fig:attention}, we feed into the models an aligned pair whose caption is ``young woman seated on the beach", we visualize the local cross-modality alignment between regions and tokens.
we found our full model \ModelName~can better attend on the described regions, showing higher-quality alignment is learned through the proposed pre-training.
More visualizations are in the supplementary file.

\section{Conclusion}
We propose an unsupervised vision-and-language pre-training approach via retrieval-based multi-granular alignment to learn strong vision and language joint representations with un-aligned text and image sources. We introduce two core designs of our proposed approach: (1) construct a retrieval-based weakly-aligned image-text corpus. (2) multi-granular pre-training objectives to enable the model to capture the cross-modal alignment at different granularity levels. 
Our experiments show that our model can consistently outperform the previous state-of-the-art unsupervised pre-trained models and can achieve similar performance as the fully-aligned pre-trained models. 
\vspace{-0.4cm}
\paragraph{Limitations:} 
As we only consider the detected object list to retrieve the candidate sentences, the retrieved sentences often do not cover other visually grounded information compared to the ground truth captions.
Besides, the detected object tags are often those general concepts lacking diversity. 
% For example, in row 3 in Fig.~\ref{fig:visualization}, the stadium in the image is detected as ``building", which leads to the failure of retrieving any candidate that covers this object. Another limitation is that each object in the object list contributes equally during the retrieval. 
% This is sub-optimal as different objects should be weighted differently based on their detection confidence and visual importance. 
% Otherwise, the retrieved candidate sentences might focus on less important objects. For example in row 3, the retrieved sentences only cover ``tree" or ``water" from the object list, which leads to semantic discrepancy between the retrieved sentences and the image. 
Our retrieval results and in turn our pre-trained models could be affected by such limitations.
We hope to address the issue by learning a Siamese network between visual concepts and sentence for better retrieval and exploiting even larger uni-modal datasets to increase the diversity in the future research.
\vspace{-0.5cm}
\paragraph{Societal Impact:} 
The models are trained on the public datasets widely used in the community.
However, these datasets are known with biases, which may in turn affect our model predictions.
We do not recommend relying on the models to make real-world decisions.
%%%%%%%%% REFERENCES
{\small
\bibliographystyle{ieee_fullname}
\bibliography{egbib}
}

%%%%%%%%% Appendix
\clearpage
\appendix

\section{Details of Motivation Study}
As introduced in Section~\ref{section:intro}, we try to answer two questions: $(i)$ whether presenting a joint image-text data from non-parallel sources would improve the learned joint embedding space than alternatively presenting uni-modal data during pre-training. $(ii)$ If we fed joint image-text data to the model, how does its existing latent alignment affect the cross-modal representation learning. 

We conduct the unsupervised vision and language pre-training on Conceptual Captions (CC) by shuffling the image-text pairs. 
For pre-training objectives, we apply standard MLM + MRM. 
All other pre-training setup is the same as introduced in Section~\ref{sec:training_setup}. 
We first compare the round-robin and joint MLM + MRM pre-training, whose results are shown in Table~\ref{tab:data-fedding}.
We then evaluate how the alignment degree of the pre-training dataset affects the model performance, where the degree is controlled by the ratio of originally aligned image-text data in Conceptual Captions.
Table~\ref{tab:paired-ratio} shows the detailed results of each downstream task.
Their Meta-Ave scores are also plotted in Fig.~\ref{fig:intro}.
From these results, we obtained two important messages: 
$(i)$ joint image-and-text input is more optimal for UVLP than alternatively presenting uni-modal data from unparallel image and text corpus. 
$(ii)$ The more the latent semantic alignment exists in the image-text data the better the pre-trained model performs. 

We further explore the realistic unsupervised V+L pre-training, where the images and texts are from two different sources.
Specifically, we sample the images from Conceptual Captions and the texts from Book Corpus respectively.
Table~\ref{tab:bc_alignment} shows that the pre-trained model on our weakly aligned CC image and BC sentence corpus far outperforms that on random pairs, indicating it also holds that better latent image-text alignment leads to better pre-trained model's performance under realistic setting.
\begin{table}[!h]
\centering
\small
\tablestyle{5pt}{0.80}
\begin{tabular}{l|ccccc}\toprule
\multirow{2}{*}{} &VQA2 &NLVR2 &VE & RefCOCO+ & \multirow{2}{*}{ Meta-Ave } \\
&Test-Dev &Test-P &Test &Devs & \\\cmidrule{1-6}
random &70.3 &51.2 &75.3 &76.5 & 68.3 \\
proposed & \bf 71.2 & \bf 67.1 & \bf 77.1 & \bf 79.7 & \bf 73.8 \\
\bottomrule
\end{tabular}
\vspace{-0.3cm}
\caption{Pre-training on realistic CC + BC data}
\label{tab:bc_alignment}
\end{table}

\section{Effectiveness of Weighted ITM}
We compared the performance of pre-training our model with or without weighted ITM. 
The models are pre-trained on CC images and texts. 
As shown in Table~\ref{tab:WITM}, weighted ITM are consistently better than treating all the retrieved pairs with the same weight.

\begin{table}[!htp]\centering
\footnotesize
\tablestyle{3pt}{0.80}
\begin{tabular}{l|ccccc}\toprule
\multirow{2}{*}{} &VQA2 &NLVR2 &VE & RefCOCO+ & \multirow{2}{*}{ Meta-Ave } \\
&Test-Dev &Test-P &Test &Devs & \\\cmidrule{1-6}
w/o $w_{\text{ITM}}$ &71.9 &72.6 &77.0 &79.7 & 75.3 \\
$w_{\text{ITM}}$ & \bf 72.1 & \bf 73.4 & \bf 77.3 & \bf 80.3 & \bf 75.8 \\
\bottomrule
\end{tabular}
\vspace{-0.3cm}
\caption{Ablation Study on weighted ITM}
\label{tab:WITM}
\end{table}

\begin{table*}[!ht]\centering
\small
\begin{tabular}{l|c|c|c|ccc|c}\toprule
\multirow{2}{*}{ Pre-training } &VQA2 &NLVR2 &VE & \multicolumn{3}{c|}{RefCOCO+} & \multirow{2}{*}{ Meta-Ave } \\
&Test-Dev &Test-P &Test &Dev &TestA &TestB & \\\cmidrule{1-8}
Round-Robin MLM+MRM &70.4 &51.1 &74.8 &73.3 &78.3 &\textbf{67.4} & 67.4 \\
Joint MLM+MRM &\textbf{70.6} &\textbf{52.4} &\textbf{74.9} &\textbf{74.5} &\textbf{79.4} &66.8 & \textbf{68.1} \\
\bottomrule
\end{tabular}
\vspace{-0.3cm}
\caption{Detailed evaluation results on four V+L downstream tasks with two different data feeding strategy for UVLP: (1) joint image-text data (joint MLM+MRM); (2) alternative uni-modal data (round-robin MLM+MRM).}
\label{tab:data-fedding}
\end{table*}

\begin{table*}[!ht]\centering
\small
\begin{tabular}{l|c|c|c|ccc|c}\toprule
\multirow{2}{*}{ Paired Ratio } &VQA2 &NLVR2 &VE & \multicolumn{3}{c|}{RefCOCO+} & \multirow{2}{*}{ Meta-Ave } \\
&Test-Dev &Test-P &Test &Dev &TestA &TestB & \\\cmidrule{1-8}
0\% &70.6 &52.4 &74.9 &74.5 &79.4 &66.8 & 68.1 \\
20\% &71.1 &70.0 &76.4 & 76.3 &80.3 &67.5 & 73.5 \\
40\% &71.4 &71.6 &77.2 &77.9 &82.4 &68.8 & 74.5 \\
60\% &71.9 &74.5 &77.8 &79.9 &84.4 &69.9 & 76.0 \\
80\% &72.2 &75.7 &78.4 &80.9 &85.7 &71.8 & 76.8 \\
100\% &72.5 &75.9 &78.7 &82.1 &86.6 &75.0 & 77.3 \\
\bottomrule
\end{tabular}
\vspace{-0.3cm}
\caption{Detailed evaluation results on four V+L downstream tasks with 6 sets of image and text corpus of different latent cross-modal alignment degree. The alignment degree is controlled by changing the ratio of original aligned image-text data from 0\% to 100\%.}
\label{tab:paired-ratio}
\end{table*}

\begin{figure*}[h!]
\centering
\includegraphics[width=14cm]{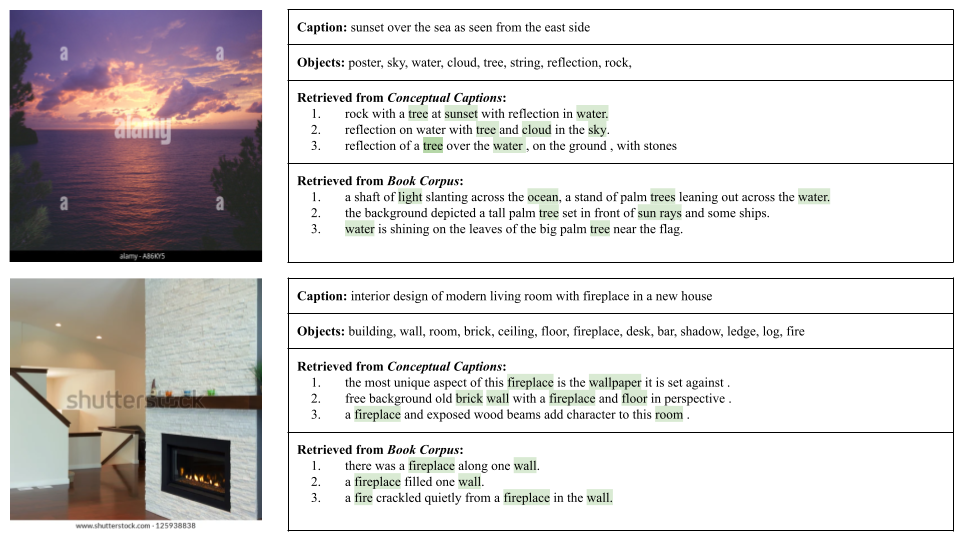}
\vspace{-0.3cm}
\caption{Examples of retrieved text from both CC and BC. The covered grounded noun phrases in retrieved sentences are highlighted in green bar for positive examples.}
\label{fig:pos-ret}
\end{figure*}

\section{Downstream Task Details}
We describe the details of fine-tuning on the four different downstream tasks: Visual Question Answering (VQA2), Natural Language for Visual Reasoning (NLVR2), Visual Entailment (VE), and Referring Expression (RefCOCO+). We mainly follow the setup of UNITER\cite{chen2020uniter} for each downstream task with minor adjustments.  

\noindent\textbf{VQA2}
Given a question about an image, the task is to predict the answer to the question. Following \cite{yu2019mcan}, we take 3,129 most frequent answers as answer candidates. We use both training and validation sets from VQA 2.0 for fine-tuning. Following UNITER, we also leverage data from Visual Genome\cite{krishna2017visualgenome} to augment the best performance on the test-dev split. We fine-tune the model with a binary cross-entropy loss with a peak learning rate of $6\times10^{-5}$ for 20 epochs. The training batch size is set as 480. 

\noindent\textbf{NLVR2}
NLVR2 is a task for visual reasoning. The objective is to determine whether a natural language statement is true or not given a pair of input images. 
We follow UNITER's setup treating each data point as two text-image pairs with repeated text. 
The two [CLS] outputs from the model are then concatenated as the joint embedding for the example. We apply a multi-layer perceptron (MLP) classifier on top of this joint embedding for the final classification. Unlike~\cite{li2020unsupervised} that conducts additional ``pre-training" on NLVR2 datasets, we only fine-tune the model with the task-specific objective to maintain the same setting as all other downstream tasks. We train the model for 8 epochs with a batch size of 60 and a peak learning rate of $3\times10^{-5}$. 

\noindent\textbf{VE}
Visual Entailment is a task built on Flickr30k Images\cite{young-etal-2014-image}, where the goal is to determine the logical relationship between a natural language statement and an image. Similar to the Natural Language Inference problem in NLP, this task is formatted as a 3-way classification problem to predict if the language statement entails, contradicts, or is undetermined with respect to the given image. An MLP transformer classifier is applied to the output of the $\text{[CLS]}$ token to make the final prediction. The model is fine-tuned using cross-entropy loss. We set the batch size as 480 and the peak learning rate as $8\times10^{-5}$. The model is fine-tuned for 4 epochs for this downstream task. 

\noindent\textbf{RefCOCO+}
The referring expression task involves locating an image region given a natural language phrase. We use RefCOCO+ \cite{yu2016modeling} as the evaluation dataset. Bounding box proposals from VinVL object detectors are used for fine-tuning. A proposal box is considered correct if it has an IoU with a gold box larger than 0.5. We add an MLP layer on top of the region outputs from the Transformer to compute the alignment score between the language phrase and each proposed region. We fine-tune our model for 20 epochs with a peak-learning rate of $2\times10^{-4}$.

\begin{figure*}[h!]
\centering
\includegraphics[width=14cm]{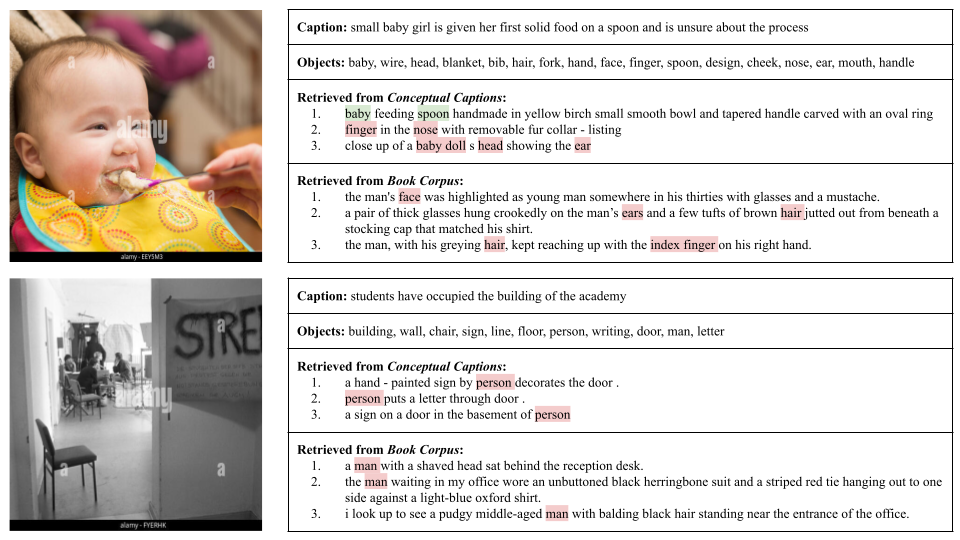}
\vspace{-0.1cm}
\caption{Examples of retrieved text from both CC and BC. The mistakenly covered grounded noun phrases in retrieved sentences are highlighted in red bar for negative examples.}
\label{fig:neg-ret}
\end{figure*}

\begin{figure}[h!]
\centering
\includegraphics[width=0.7\linewidth]{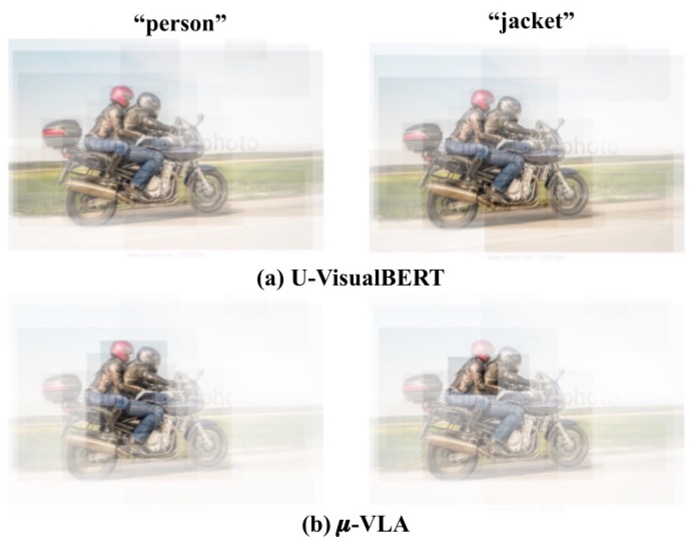}
\vspace{-0.2cm}
\caption{Text-to-image attention given the aligned pair whose caption is ``person in a leather jacket riding a motorcycle on the road".}
\label{fig:attn_viz_1}
\end{figure}

\begin{figure}[h!]
\centering
\includegraphics[width=0.7\linewidth]{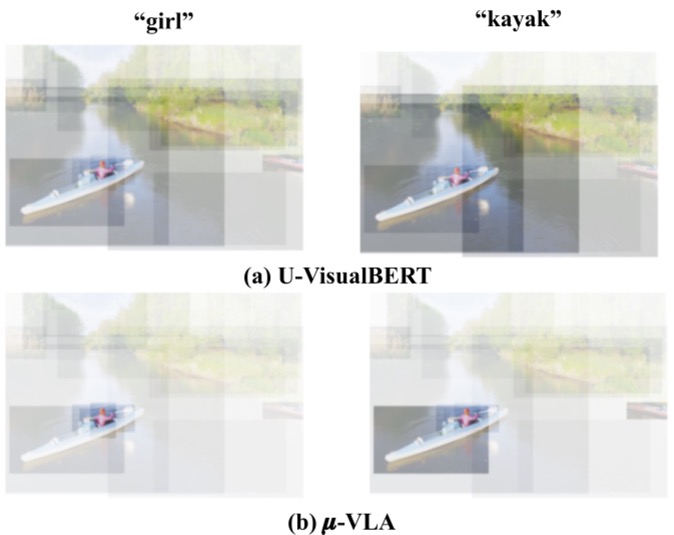}
\vspace{-0.2cm}
\caption{Text-to-image attention given the aligned pair whose caption is ``girl in a blue kayak floating on the picturesque river at sunset".}
\label{fig:attn_viz_2}
\end{figure}

\begin{figure}[h!]
\centering
\includegraphics[width=0.7\linewidth]{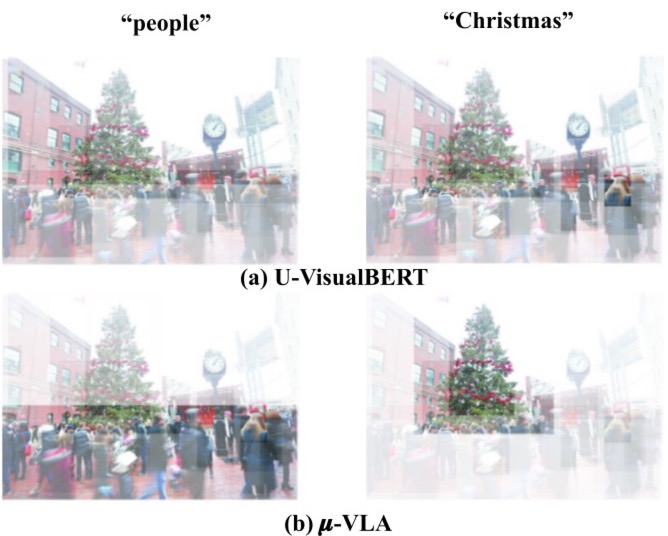}
\vspace{-0.2cm}
\caption{Text-to-image attention given the aligned pair whose caption is ``people walking by the christmas tree and stage area".}
\label{fig:attn_viz_3}
\end{figure}

\begin{figure}[h!]
\centering
\includegraphics[width=0.7\linewidth]{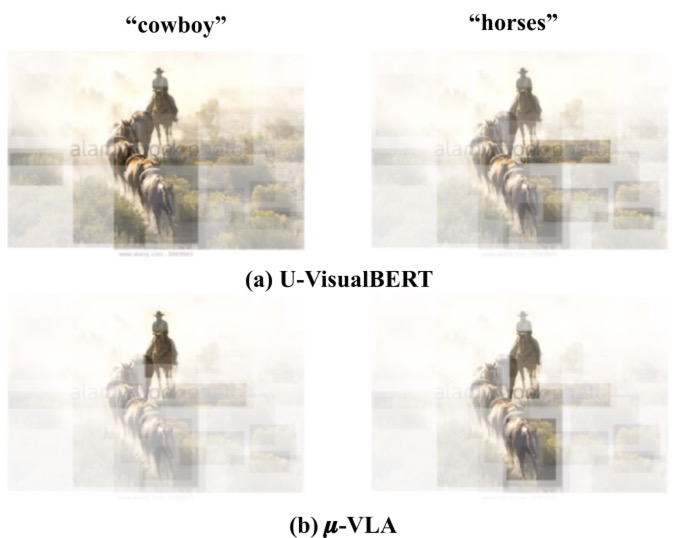}
\vspace{-0.2cm}
\caption{Text-to-image attention given the aligned pair whose caption is ``single cowboy guiding a line of horses through the desert".}
\label{fig:attn_viz_4}
\end{figure}
\section{Additional Visualization}
We present additional examples of retrieved text from both CC and BookCorpus. Specifically, we demonstrate more positive examples in Fig \ref{fig:pos-ret} that covers the appropriate grounded noun phrases. We also share some negative examples in Fig \ref{fig:pos-ret}. As analyzed in the limitation section, the current language embedding model weighs all the object tags equally to generate the joint embedding representation. This can lead to mistakenly focused object tags when retrieving the text. In row 1 of Fig \ref{fig:neg-ret}, texts retrieved cover less important noun phrases such as ``finger" and ``hair" instead of the more important noun phrase "baby". Row 2 of Fig \ref{fig:neg-ret} demonstrate mistakenly retrieved texts due to the limitation of the pre-defined object categories in the object detector. In this example,  the students in the image are detected as ``person" or ``man", which leads to the failure of retrieving any valid text.    

We also demonstrate more examples on text-to-image attention between the pre-trained U-VisualBert and {\ModelName } on the Conceptual Captions Validation set in Fig \ref{fig:attn_viz_1}, \ref{fig:attn_viz_2}, \ref{fig:attn_viz_3}, \ref{fig:attn_viz_4}. These examples provide additional evidence on the better local alignment captured by \ModelName. 

\end{document}